\documentclass[letterpaper, 10 pt, conference]{ieeeconf}  % Comment this line out
\IEEEoverridecommandlockouts             % This command is only
\usepackage{graphicx}
\usepackage{url}
\usepackage{slashbox}
\usepackage{amsmath, amssymb}
\usepackage{algorithmic}
\usepackage{algorithm}

\title{\LARGE \bf
Online Semantic Exploration of Indoor Maps}

\author{\parbox{5 in}{\centering Ziyuan Liu, Dong Chen and Georg von Wichert
\thanks{Z. Liu, and D. Chen are with the Institute of Automatic Control Engineering, Technische Universit\"at M\"unchen, D-80290, M\"unchen, Germany. \small{\texttt{ziyuan.liu@tum.de}},\small{\texttt{chendong@mytum.de}} }
\thanks{G. von Wichert is with Siemens AG, Corporate Research \& Technologies, Munich, Germany and Institute for Advanced Study, Techniche Universit\"at M\"unchen, Munich, Germany \small{\texttt{georg.wichert@siemens.com}}  }
}
}

\begin{document}

\maketitle
\thispagestyle{empty}
\pagestyle{empty}

%%%%%%%%%%%%%%%%%%%%%%%%%%%%%%%%%%%%%%%%%%%%%%%%%%%%%%%%%%%%%%%%%%%%%%%%%%%%%%%%
\begin{abstract}
%The primary challenge for any autonomous system operating in realistic, rather unconstrained scenarios is to manage the complexity and uncertainty of the real world. 
%While it is unclear how exactly humans and other higher animals master these problems, it seems evident, that abstraction plays an important role. 
%The use of abstract concepts allows to define the system behavior on higher levels. Abstractly defined behavior will provide guidance in a wide range of situations and thus increase the overall robustness of the system. 
In this paper we propose a method to extract an abstracted floor plan from typical grid maps using Bayesian reasoning. The result of this procedure is a probabilistic generative model of the environment defined over abstract concepts. It is well suited for higher-level reasoning and communication purposes. We demonstrate the effectiveness of the approach through real-world experiments.
\end{abstract}

%%%%%%%%%%%%%%%%%%%%%%%%%%%%%%%%%%%%%%%%%%%%%%%%%%%%%%%%%%%%%%%%%%%%%%%%%%%%%%%%
\section{Introduction and related work}
Most of todays' mapping approaches aim to construct a globally consistent, metric map of the robot's operating environments. See Fig.~\ref{figure:exp-final} a) for a typical result. Such maps enable the robot to localize itself with respect to the environment. Based on this capability, the robot can also plan a path and navigate towards a goal, that will be specified by its metric position in the global map reference frame. However, the robots do not understand their environment in terms of typical semantic concepts like rooms, corridors or even functionally enriched concepts like kitchen or living room. Furthermore, the robots do not understand relations like adjacency, connectivity via doors, or properties like rectangularity that -- if known to be relevant to the given environment -- could help to build the maps in the first place.

Assigning semantics to spatial maps in robotics has not been looked at as intensely as metric or topological mapping. Still, several important contributions to the field have already been made. They can be clustered into two major groups. The first group consists of methods based on place labeling, some notable examples are \cite{wolf2008semantic, pronobis2010cogsys, friedman2007voronoi,goerke2009building,mozos2006supervised}. These methods assign semantic labels to places or regions of the accessible work space of the robot. They are very much in the tradition of \cite{thrun1996integrating}.

A second group is formed by approaches assigning semantic labels to parts or objects of the perceived structure of the environment, like traversable terrain, trees or similar structures in outdoor environments or walls, ceilings, and doors in indoor settings~\cite{limketkai2005relational, douillard2008laser, nuechter2008towards,tong2010,wang2011,krishnan2010,persson2007}. 

%%%%%%%%%%%%%%%%%%%%%%%%%%%%%%%%%%%%%%%%%%%%%%%%%%%%%%%%%%%%%%%%%%%%%%%%%%%%%%%%
\section{Problem description}

\begin{figure*}[!htb]
\centering
\includegraphics[width=.95\textwidth]{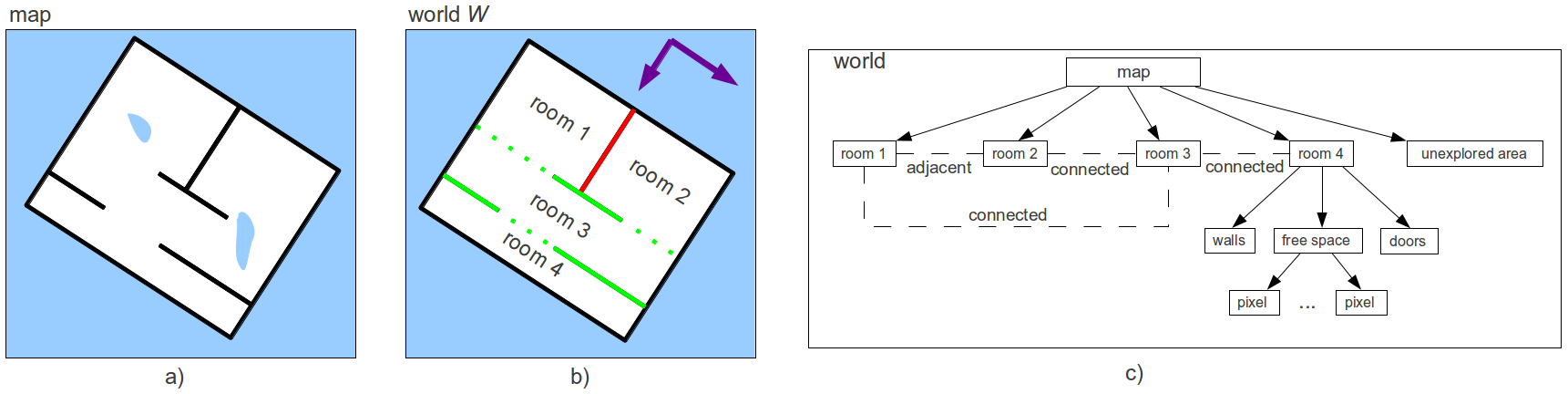}
\caption{a) A simplified occupancy grid map: Unexplored area is drawn in blue, free space is drawn in white. Occupied area is drawn in black. b) A possible floor plan represented as a scene graph ($W$): The world is divided into four rooms and the corresponding unexplored area. Connectivity is given by the color of walls:
the color green indicates \emph{connected}, which means there is a door (green dotted) between two rooms; the color red means \emph{adjacent}, which means that two rooms are neighbor and do not connect themselves through a door; the color black stands for a boundary wall. The detected main orientations are illustrated by violet arrows. c) The semantic description of the world in form of the scene graph: Directed links connect nodes. The dashed lines represent connectivity. Like room 4, each room has three child nodes: walls, free space, and doors. Note that the lowest level of node in the tree structure is the image pixel that belongs to walls, free space and doors.}
\label{figure:example}
\end{figure*}

Different from those methods mentioned above, we aim to construct a probabilistic generative model of the world around the robot, that is essentially based on abstract semantic concepts but at the same time allows to predict the continuous percepts that the robot obtains via its noisy sensors. This abstract model has a form similar to a scene graph, a structure which is widely used in computer graphics. The graph (see Fig.~\ref{figure:example} c) in our case consists of rooms and doorways connecting the rooms and can be visualized as a classical floor plan(see Fig.~\ref{figure:example} b).

The scene graph and the semantically annotated world state are represented as a vector of parameters $W$ observed through the occupancy map $M$. In Bayesian framework we can use a maximum posterior approach to infer the most probable state $W^*\in\Omega$ from the space of possible worlds $\Omega$ given the map $M$. 

\begin{equation}
W^*=\arg\!\max_{\!\!\!\!\!\!\!\!\!\!\!_{W\in\Omega}} \, p(W|M),
\label{eq:argmax}
\end{equation}
where
\begin{equation}
p(W|M)\propto p(M|W)p(W).
\label{eq:generative}
\end{equation}

Here $p(W|M)$ is the posterior distribution of $W$ given a map $M$, and $p(W)$ is the prior specifying, which worlds $W$ are possible at all. $p(M|W)$ is the likelihood function describing how probable the observed map $M$ is, given the different possible worlds represented by a parameter vector $W$. The actual semantic model is represented in the structure of the parameter vector $W$, while semantically relevant constraints go into the prior $p(W)$. 

\section{A generative model for occupancy grids}

The prior $p(W)$ in (\ref{eq:generative}) expresses a set of assumptions concerning the structured world based on context knowledge as follows:
\begin{enumerate}
	\item[1)] a room has four walls and possesses a rectangular shape.
	\item[2)] a room has at least one door, and a door is placed on a wall.
	\item[3)] each cell in the map should only belong to one room.
	\item[4)] walls of an indoor environment have two main orientations (see Fig. \ref{figure:example} b)).
\end{enumerate}
The prior $p(W)$ penalizes worlds that are not fully compliant with the above assumptions:
\begin{equation}
p(W)=\alpha_1\times\alpha_2\times\alpha_3\times\alpha_4,
\end{equation}
where $\alpha_1$, $\alpha_2$, $\alpha_3$ and $\alpha_4$ are the corresponding penalization terms for the point 1), 2), 3) and 4) of the prior information respectively.  $\alpha_1$, $\alpha_2$,  and $\alpha_4$ are set to 
 $\psi_1, \psi_2, \psi_4\in(0,1)$ if there is a conflict with the corresponding constraint, and $\alpha_3$ is 

\begin{eqnarray}
\alpha_3&=&\prod\limits_{c(x,y)\in M}\psi_3^{\gamma(c(x,y))},\nonumber\\
\gamma(c(x,y))&=&\left\{\begin{array}{lc}
\sigma(c(x,y))-1,\sigma(c(x,y))>1,\\
0,\textrm{otherwise},\\
\end{array}
\right.
\end{eqnarray}
where $\psi_3$ satisfies $\psi_3 \in(0,1)$. $c(x,y)$ denotes one grid cell in the map $M$. $\sigma(c(x,y))$ indicates the number of rooms, to which $c(x,y)$ belongs. $\alpha_3$ is a cell-wise penalization of the overlap between different rooms.  

Furthermore, for our generative model, we need to specify the likelihood function $p(M|W)$. Since $M$ is represented by an occupancy grid with statistically independent grid cells $c \in M$, we only need to come up with a model $p(c|W)$ for all cells at their locations $(x,y)$ in the map M:

\begin{equation}
p(M|W) = \prod_{c(x,y) \in M} p(c(x,y)|W).
\label{eq:prod}
\end{equation}

For our model $p(c(x,y)|W)$, we first discretize the cell state $M(x,y)$ by classifying the intensity values into three classes $C_M(x,y)$ according to:

\begin{equation}\label{equ:classify}
C_M(x,y)=\left\{\begin{array}{lcc}
2,\quad 0\leq M(x,y)\leq h_o,\\
1, \quad h_o<M(x,y)\leq h_u,\\
0,\quad h_u<M(x,y)\leq h_f,
\end{array}
\right.
\end{equation}
where $h_o$, $h_u$ and $h_f$ are the intensity thresholds for occupied, unexplored and free pixels, respectively. Ideally, these parameters should be learned from training data, so that they do not largely depend on the input data. Based on our world model $W$ we can also predict expected cell states $C_W(x,y)$ accordingly:

\begin{equation}
C_W(x,y)=\left\{\begin{array}{lcc}
2,\quad (x,y)\in S_w,\\
1,\quad (x,y)\in S_u,\\
0,\quad (x,y)\in S_f,
\end{array}
\right.
\end{equation}
where $S_w, S_u$ and $S_f$ are the set of all the wall pixels, unknown pixels and free space pixels in the world $W$ respectively. $p(c(x,y)|W)$ can then be represented in the form of a lookup-table.

\begin{table}[htb]
	\centering
	\begin{tabular}{|c||*{3}{c|}}\hline
	\backslashbox{$C_W(x,y)$}{$C_M(x,y)$}
	&\makebox[3em]{0}&\makebox[3em]{1}&\makebox[3em]{2}\\\hline\hline
	%%&0&1&2\\\hline\hline
	0 (wall)&0.5&0.1&0.1\\\hline
	1 (unknown)&0.3&0.8&0.1\\\hline
	2 (free space)&0.2&0.1&0.8\\\hline
	\end{tabular}
	\caption{The lookup table for $p(c(x,y)|W)$.}
	\label{TAB:mapping table}
\end{table}

In principle the likelihood $p(c(x,y)|W)$ plays the role of a sensor model. In our case it captures the quality of the original mapping algorithm producing the grid map (including the sensor models for the sensors used during the SLAM process), and could be learned from labeled training data. However, for the experiments described in section~\ref{exp} we used the manually selected values given in Table \ref{TAB:mapping table}.

%%%%%%%%%%%%%%%%%%%%%%%%%%%%%%%%%%%%%%%%%%%%%%%%%%%%%%%%%%%%%%%%%%%%%%%%%%%%%%%%
\section{Searching the solution space}
For solving (\ref{eq:argmax}) we need to efficiently search the large and complexly structured solution space $\Omega$. Here we adopt the approach of \cite{zhu2000integrating}, who propose a data driven Markov chain Monte Carlo (MCMC) technique for this purpose. The basic idea is to construct a Markov Chain that generates samples $W_i$ from the solution space $\Omega$ according to the distribution $p(W|M)$ after some initial burn-in time. One popular approach to construct such a Markov chain is the Metropolis-Hastings (MH) algorithm \cite{metropolis53,hastings70}. In MCMC techniques the Markov chain is constructed by sequentially executing state transitions (in our case from a given world state $W$ to another state $W'$) according to a transition distribution $\Phi(W'|W)$ of the sub-kernels. An example of $\Phi(W'|W)$ is given in Table \ref{TAB:Transition}. In order for the chain to converge to a given distribution, it has to be reversible and ergodic~\cite{bishop2007machine}. The MH algorithm achieves this by generating new samples in three steps. First a transition is proposed according to $\Phi(W'|W)$, subsequently a new sample $W'$ is generated by a proposal distribution $Q(W'|W)$, and then it is accepted with the probability $\lambda$.

\begin{equation}
\lambda(W,W') = \min\left( 1, \frac{p(W'|M) Q(W|W')}{p(W|M) Q(W'|W)} \right)
\label{eq:MH}
\end{equation}

The resulting Markov chain can be shown to converge to $p(W|M)$. However, the selection of the proposal distribution is crucial for the convergence rate. Here, we follow the approach of \cite{zhu2000integrating} to propose state transitions for the Markov chain using discriminative methods for the bottom-up detection of relevant environmental features (e.g. walls, doorways) and construct the proposals based on these detection results. An overview of our online semantic exploration algorithm is described in Algorithm \ref{alg0}.

\begin{algorithm}
%\small
\footnotesize
\caption{Online Semantic Exploration}
\label{alg0}
\begin{algorithmic}
\REQUIRE{input map $M$}
%\STATE{Generate $M$ hypotheses for estimating size parameters:}
%\STATE{3-step hierarchical estimation of subsets:}
    \WHILE{certain accuracy condition not satisfied}
       	\IF{input map updated}
       	    \STATE{generate the classified map $C_M(x,y)$;}
        	\STATE{generate new room candidates;}
		\ENDIF
        \STATE{select one sub-kernel according to the transition probabilities $\Phi(W'|W)$;}
        \STATE{calculate the acceptance probability $\lambda(W,W')$ of the selected sub-kernel according to the MH algorithm (\ref{eq:MH});}
        \STATE{draw a random float number $\theta$, $\theta\in\mathcal{U}[0,1)$;}
        \IF{$\theta<\lambda(W,W')$}
       	    \STATE{accept;}
       	\ELSE
        	\STATE{reject;}
		\ENDIF
        %\STATE{Generate $M$ new hypotheses of corresponding subset:}
    \ENDWHILE
\end{algorithmic}
%\end{small}
\end{algorithm}

\subsection{MCMC Kernels}
In order to realize the Markov chain in form of the Metropolis-Hastings algorithm, we arrange the kernels that alter the structure of the world as reversible pairs:
\begin{itemize}
	\item Kernel pair 1: ADD or REMOVE one room.
		%\begin{itemize}
			%\item ADD:  draw one new room from certain candidates, then try to add this room to the world.
			%\item REMOVE: try to cancel one existing room from the world.
	%	\end{itemize}
	\item Kernel pair 2: SPLIT one room or MERGE two rooms.
	%	\begin{itemize}
			%\item SPLIT: try to decompose one existing room into two rooms.
			%\item MERGE: try to combine two existing rooms, and generate one new room out of them.
		%\end{itemize}
	\item Kernel pair 3: SHRINK or DILATE one room.
		%\begin{itemize}
			%\item SHRINK: try to move one wall of one room along certain orientation, so that the room becomes smaller.
			%\item DILATE: similarly to SHRINK, move one wall of one room, so that the room becomes bigger.
	%	\end{itemize}
	\item Kernel pair 4: ALLOCATE or DELETE one door
	%	\begin{itemize}
			%\item ALLOCATE: draw one door from the door candidates that are provided by door detector, then try to assign it to two existing rooms.
			%\item DELETE: cancel one assigned door.
		%\end{itemize}
\end{itemize}

\begin{figure}[htb]
	\centering
  	\includegraphics[width=0.95\columnwidth]{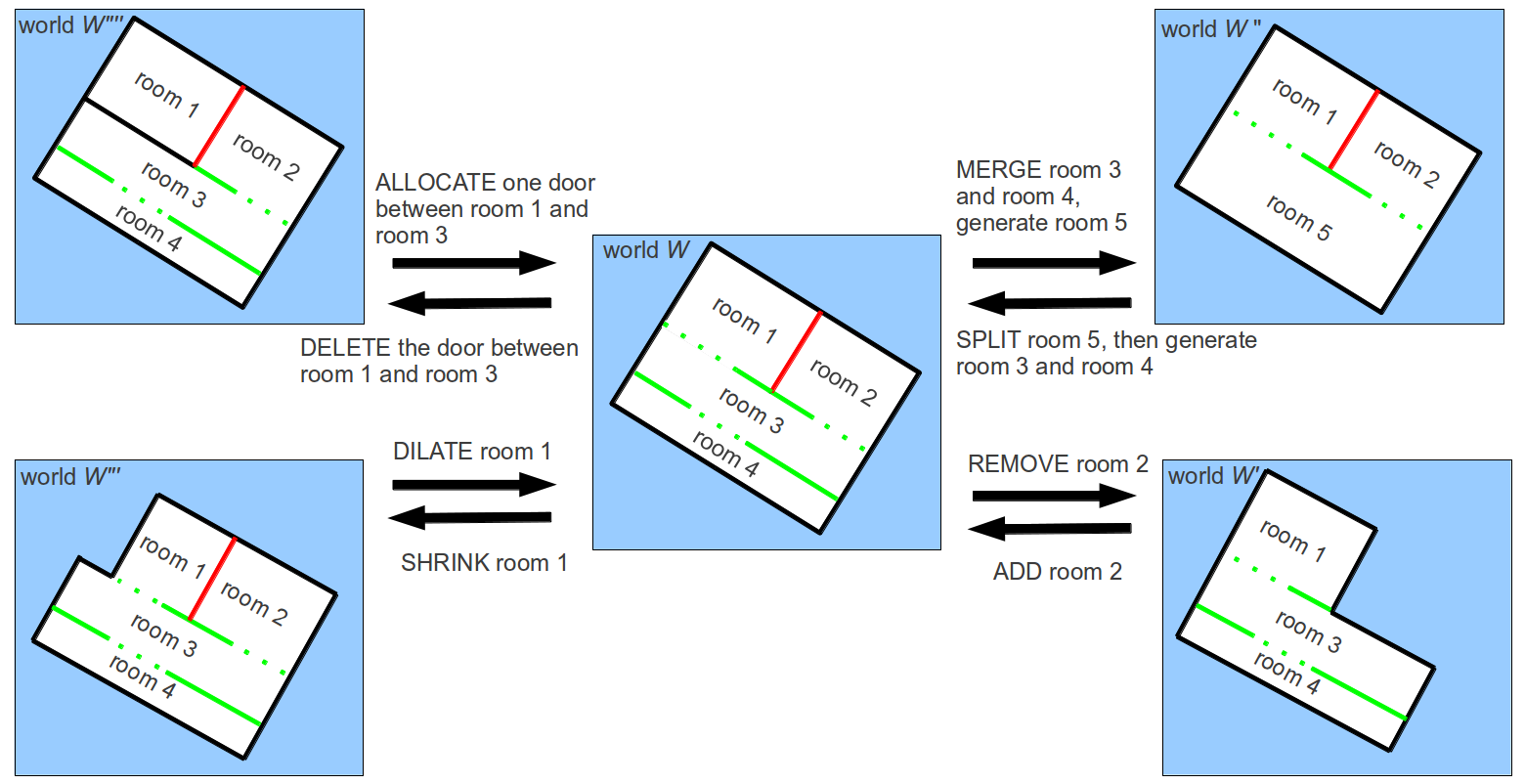}
    \caption{Reversible MCMC kernel pairs: ADD/REMOVE, SPLIT/MERGE, SHRINK/DILATE and ALLOCATE/DELETE.}
   	\label{figure:mcmc-kernels}
\end{figure}

Fig. \ref{figure:mcmc-kernels} shows an example of the four reversible MCMC kernel pairs. The world $W$ can transit to $W^{'}$, $W^{''}$, $W^{'''}$ and $W^{''''}$ by applying the sub-kernel REMOVE, MERGE, SHRINK and DELETE, respectively. By contrast, the world $W^{'}$, $W^{''}$, $W^{'''}$ and $W^{''''}$ can also transit back to $W$ using corresponding reverse sub-kernels. In the following, we discuss the kernels in detail.

\subsubsection{ADD}
The sub-kernel ADD tries to put one new room into the world. Once the input map is updated, several room candidates are generated around the current robot position, then one room is selected from the generated rooms in a resampling style \cite{kitagawa}. Each of the generated room candidates are weighted according to how well their walls match the observations provided by the occupancy grid map. The weight of a room $\omega_r$ is defined as the lowest wall weight $\omega_{w_{j}}$ among its four walls, where $j,j\in\{r_1,r_2,r_3,r_4\}$, indexes the wall, with $r_i,i\in\{1,2,3,4\}$, indicating the $i$th wall of room $r$:

\begin{equation}
\omega_r=\min_{j\in\{r_1,r_2,r_3,r_4\}}{\omega_{w_{j}}}.\label{equ:roomweight}
\end{equation}

The wall weight $\omega_{w_{ j}}$is calculated as:

\begin{equation}
\omega_{w_{ j}}=\frac{n(w_{ j})}{l(w_{ j})},\label{equ:wallweight}
\end{equation}
where $l(w_{ j})$ indicates the length of wall $w_{ j}$ and can be computed from the coordinates of its two end points $(x_{w_{ j,1}},y_{w_{ j,1}}),(x_{w_{ j,2}},y_{w_{ j,2}})$:

\begin{equation}
l(w_{ j})=\sqrt{(x_{w_{ j,1}}-x_{w_{ j,2}})^2+(y_{w_{ j,1}}-y_{w_{ j,2}})^2}.\label{equ:length}
\end{equation}

The term $n(w_{ j})$ counts the number of wall pixels that match with the map:

\begin{equation}
n(w_{ j})=\sum\limits_{(x,y)\in w_{ j}}t(x,y),\label{equ:number}
\end{equation}

where

\begin{equation}\label{equ:01}
t(x,y)=\left\{\begin{array}{lc}
1,\quad C_{M}(x,y)=0,\\
0,\quad \textrm{otherwise}.
\end{array}
\right.
\end{equation}
The normalized weights $\omega{'}_r$ is calculated as:

\begin{equation}
\omega^{'}_r=\frac{\omega_r}{\sum\limits_{r\in B}\omega_r},\label{equ:normalize}
\end{equation}
where $B$ indicates the set of all the room candidates generated using the robot position. Then, we calculate the cumulative weights $A_r$ for room $r$:

\begin{equation}
A_r=\sum\limits_{i=1}^r\omega_i.
\end{equation}

Finally, we can draw a room candidate $n$ out of $B$, by generating a random number $k,k\in[0,1)$,

\begin{equation}
n=\min\{i|k\leq A_i\}.\label{equ:sampling}
\end{equation}

\begin{figure}[htb]
	\centering
  	\includegraphics[width=0.95\columnwidth]{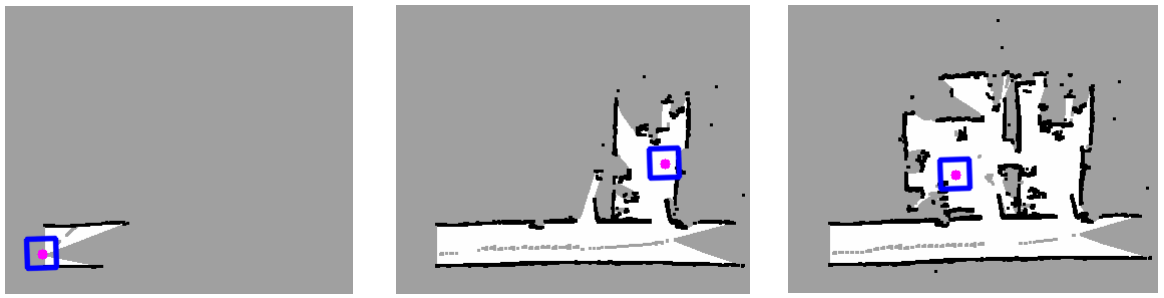}
    \caption{Three examples of generating only one room candidate of certain minimum size using robot position as center. Blue rectangles show the generated rooms, and the violet points show robot position.}
   	\label{figure:add}
\end{figure}

There are many ways to generate room candidates using robot position. Fig. \ref{figure:add} depicts one of the simplest ways of generating room candidates, which generates just one room of certain minimum size using robot position as the center of the generated room. 

\subsubsection{SPLIT}
The sub-kernel SPLIT tries to decompose one member room into two rooms. To do this, a member room $r$ is drawn randomly from the current world $W$ according to a uniform distribution. Subsequently, we use Hough line detection in the room $r$ to find all line segments which are potential splitting possibilities. Let $E_r$ denote the set of the detected line segments within room $r$. Each detected line segment $e,e\in E_r$ is weighted, using its length $l(e)$:

\begin{equation}
\omega_e=l(e),
\end{equation}
where the length $l(e)$ is similarly calculated as done in (\ref{equ:length}). Then we normalize the weights and build the cumulative distribution of $E_r$. Furthermore, we draw one line segment out of $E_r$, as done in ((\ref{equ:normalize}) to (\ref{equ:sampling})).

The selected line segment is then extended to intersect with the walls of the room, so that two rooms are generated as the result of SPLIT. % The MH algorithm then decides whether this action is accepted. 
\subsubsection{MERGE}

The sub-kernel MERGE tries to combine two member rooms of the current world $W$. The first room $r$ is drawn from the set of all member rooms of world $W$ according to a uniform distribution. Additionally, a second room $s$ needs to be selected from the rest of the member rooms. For sampling $s$, we define a new weight $a_r(s)$, which is the reciprocal of the distance $d(c_r,c_s)$ between the center point $c_r$ of room $r$ and the center point $c_s$ of room $s$:

\begin{equation}
d(c_r,c_s)=\sqrt{(c_r.x-c_s.x)^2+(c_r.y-c_s.y)^2},
\end{equation}
where $(c_r.x,c_r.y)$ and $(c_s.x,c_s.y)$ are the image coordinates of the two center points. The weight $a_r(s)$ is calculated as follows:

\begin{equation}
a_r(s)=\frac{1}{d(c_r,c_s)}.
\end{equation}

Applying the same sampling technique as that in (\ref{equ:normalize}) to (\ref{equ:sampling}), we can obtain the second room $s$. Then we try to combine them into one room. The underlying idea for using $a_r(s)$ in the sampling is that the closer two rooms are, the more likely they can be combined.

\subsubsection{SHRINK and DILATE}

The kernel pair SHRINK/DILATE tries to move a wall $w_j$ of a member room $r$ of the current world $W$ along one of the main orientations. Here, $j,j\in\{r_1,r_2,r_3,r_4\}$, indexes the wall, with $r_i,i\in\{1,2,3,4\}$, indicating the $i$th wall of room $r$. For selecting the room $r$ from the set of all member rooms, we define a new weight $b_r$:

\begin{equation}\label{equ: new room weight}
b_r=\left\{\begin{array}{cl}
 \frac{1}{\omega_r}, &\frac{1}{\omega_r} \leq h_b\\
h_b, &\textrm{otherwise},
\end{array}
\right.
\end{equation}
where $\omega_r$ is the room weight defined in (\ref{equ:roomweight}). $h_b$ is a predefined threshold for the weight. Using $b_r$, a room is drawn according to (\ref{equ:normalize}) to (\ref{equ:sampling}). 
%The underlying idea for using this weight is that the worse a room matches the map, the more necessarily it should be changed by SHRINK/DILATE. 

Once the room is selected, one wall $w_j$ needs to be drawn from its four walls. Following the same idea, we define a new weight $v_{w_j}$ for sampling the wall:

\begin{equation}
v_{w_j}=\left\{\begin{array}{cl}
 \frac{1}{\omega_{w_j}}, &\frac{1}{\omega_{w_j}} \leq h_v\\
h_v, &\textrm{otherwise},
\end{array}
\right.
\end{equation}
where $\omega_{w_j}$ is the wall weight defined in (\ref{equ:wallweight}), and $h_v$ is a predefined threshold. $h_b$ and $h_v$ are just the upper bound for the corresponding weight, thus they do not depend on the input data. Again, the wall is drawn according to $v_{w_j}$, as done in (\ref{equ:normalize}) to (\ref{equ:sampling}). After the wall is selected, we propose to shift it along one of the main orientations using a zero-mean Gaussian distribution. In principle, the algebraic sign decides whether a SHRINK or a DILATE is proposed, e.g. if a positive sign proposes a SHRINK, then a negative sign will propose a DILATE. In general, SHRINK and DILATE sub-kernel have both 50\% chance to be proposed. 
\subsubsection{ALLOCATE}

This sub-kernel tries to assign one door to two member rooms so as to explore connectivity information of the semantic world. A door detector which is based on blob detection \cite{chang04} proposes door candidates for the sub-kernel ALLOCATE. We draw one door candidate from the set of all candidates according to their weights. Here, the weight $\omega_g$ of a door $g$ is similar to the weight of walls $\omega_{w_j}$ that is defined in (\ref{equ:wallweight}):

\begin{equation}
\omega_{g}=\frac{n^{'}(g)}{l(g)},\label{equ:door weight}
\end{equation}
where $l(g)$ is calculated the same as in (\ref{equ:length}), and $n^{'}(g)$ is computed as follows:

\begin{equation}
n^{'}(g)=\sum\limits_{(x,y)\in g}t^{'}(x,y),
\end{equation}
where
\begin{equation}
t^{'}(x,y)=\left\{\begin{array}{lc}
1,\quad C_{M}(x,y)=2,\\
0,\quad \textrm{otherwise}.
\end{array}
\right.
\end{equation}

Using the weight $\omega_g$, one door candidate is drawn from the set of all detected candidates, as done in (\ref{equ:normalize}) to (\ref{equ:sampling}). Then, the MH algorithm decides whether this door will be accepted. Here we do not detail on how to detect the door candidates.
\subsubsection{REMOVE and DELETE}

The sub-kernel REMOVE and DELETE have similar functionality, which is to cancel one existing member room and one of the assigned doors respectively. There are no special discriminative methods used for these two sub-kernels. They just draw one member from the corresponding set (existing rooms or assigned doors) and propose to cancel this member, then the MH algorithm decides whether this proposal is accepted. Following the idea that the worse a member matches the map, the more likely it should be canceled, we use the weight $b_r$ defined in (\ref{equ: new room weight}) for room sampling. Similarly, we define a new weight $z_g$ for door sampling:

\begin{equation}\label{equ: new door weight}
z_g=\left\{\begin{array}{cl}
 \frac{1}{\omega_g}, &\frac{1}{\omega_g} \leq h_g\\
h_g, &\textrm{otherwise},
\end{array}
\right.
\end{equation}
where $\omega_g$ is the door weight defined in (\ref{equ:door weight}), and $h_g$ is a predefined threshold. 
\subsection{Proposal Probability $Q(W^{'}|W)$ and $Q(W|W^{'})$}
The proposal probability $Q(W^{'}|W)$ describes how probable the world $W$ can transit to the world $W^{'}$, and by contrast, $Q(W|W^{'})$ is the probability for transiting back to the world $W$, given the world $W^{'}$. Intuitively, $Q(W^{'}|W)$ is the product of the normalized weight of the selected elements (room candidate, splitting line, wall etc.) in the corresponding MC sub-kernel defined in the previous section. For instance, in the ADD or REMOVE sub-kernel, $Q(W^{'}|W)$ is equal to the corresponding normalized weight of the selected room candidate or that of the selected member room. 
%$Q(W^{'}|W)$ in ALLOCATE and DELETE can be calculated similarly to that in ADD and REMOVE respectively. In SPLIT, $Q(W^{'}|W)$ is the product of the corresponding normalized weight of the selected member room and that of the selected splitting line. In SHRINK/DILATE, $Q(W^{'}|W)$ is product of three terms: the corresponding normalized weight of the member room, that of the selected wall and that of the generated Gaussian shift. Similarly, $Q(W^{'}|W)$ of MERGE is calculated as the product of the corresponding normalized weight of the first room and that of the second room.

Compared with $Q(W^{'}|W)$, the calculation of $Q(W|W^{'})$ is less intuitive, because the back transition is virtual and must be defined. In the example of ADD, $Q(W|W^{'})$ should perform the same function as the sub-kernel REMOVE, namely, the world $W'$ transits back to the world $W$ by canceling the room that is added in the transition from $W$ to $W'$, thus $Q(W|W^{'})$ of ADD should be the normalized weight of the added room in the sub-kernel REMOVE. $Q(W|W^{'})$ for other sub-kernels can be similarly defined.
%$Q(W|W^{'})$ of REMOVE can also be calculated as the normalized weight of the room, that is canceled in the transition from $W$ to $W'$, in the sub-kernel ADD. Analogously, $Q(W|W^{'})$ of SHRINK, DILATE, MERGE, SPLIT, ALLOCATE and DELETE can be calculated in a style similar to $Q(W^{'}|W)$ in their corresponding reverse sub-kernels. In addition, the SHRINK/DILATE pair just tries to move one wall of the selected room using a relatively small shift, thus the resulting world $W'$ is similar to $W$. For computational simplicity, we assume that $Q(W^{'}|W)$ and $Q(W|W^{'})$ are equal in the SHRINK/DILATE pair.
%%%%%%%%%%%%%%%%%%%%%%%%%%%%%%%%%%%%%%%%%%%%%%%%%%%%%%%%%%%%%%%%%%%%%%%%%%%%%%%%
\section{Experimental results}\label{exp}

In experiments, we use the gmapping package in ros.org \cite{grisetti2007improved} to provide input maps for our algorithm. The selection probabilities of the MC sub-kernels are listed in Table \ref{TAB:Transition}. The values in Table \ref{TAB:Transition} are just an example. There are many possibilities on how to choose them. The simplest one is to equally set all the probabilities. In this sense, these values are independent of the input data.

\begin{figure}[t]
	\centering
	\includegraphics[width=0.69\columnwidth]{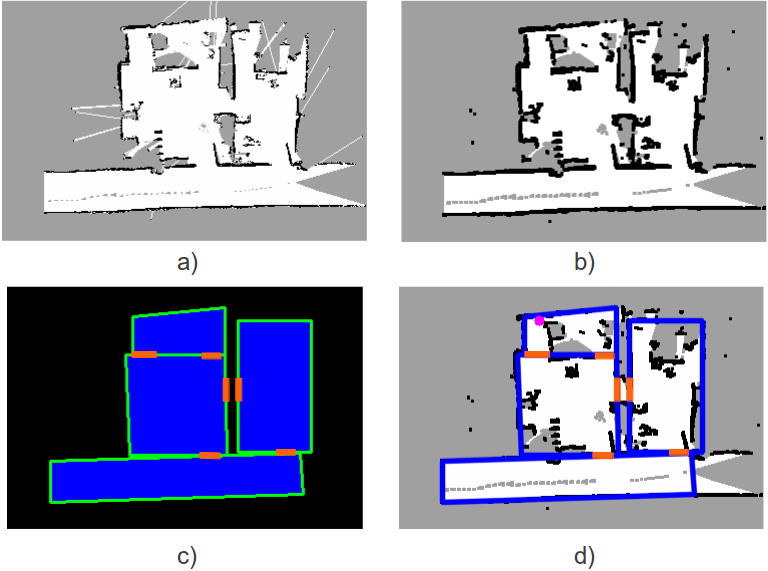}
    \caption{Final result of the semantic exploration. a) The last input map $M$ obtained from gmapping. b) The classified map $C_M(x,y)$ with three intensity values (black=wall, grey=unexplored, white=free). c) The analyzed world $W$ (green=wall, black=unknown, blue=free, orange=door). d) A direct comparison between the analyzed world and the input map: walls (blue) and doors (orange) of the world $W$ drawn into the map. Here the violet point shows the robot position.}
    \label{figure:exp-final}
\end{figure}

Fig. \ref{figure:exp-final} shows one final result of our semantic exploration algorithm. Here, part a) shows an original input occupancy map $M$ that is obtained from gmapping. Part b) shows the classified map $C_M(x,y)$ that is defined in (\ref{equ:classify}), with the color black, gray and white indicating occupied, unexplored and free pixels respectively. Part c) visualizes the world state $W$ representing our structured semantic model. Here the colors green, black, blue and orange show the wall, unknown, free and door pixels respectively. In part d), walls (blue) and doors (orange) of the world $W$ are directly plotted onto the input map $M$, so as to give a more intuitive comparison.

\begin{table}[htb]
	\caption{Transition probabilities $\Phi(W'|W)$ of MC sub-kernels.}
	\centering
	\begin{tabular}{|c||*{5}{c|}}\hline
	%\backslashbox{sub-kernel}{iteration $\beta$}&\makebox{$\beta\leq$1000}\\\hline\hline
	Sub-kernel&Probability\\\hline\hline
	ADD 	&0.2	\\\hline
	REMOVE	&0.05	\\\hline
	SPLIT	&0.125		\\\hline
	MERGE	&0.125		\\\hline
	SHRINK 	&0.2		\\\hline
	DILATE 	&0.2		\\\hline
	ALLOCATE&0.05		\\\hline
	DELETE	&0.05		\\\hline
	\end{tabular}
	\label{TAB:Transition}
\end{table}

\begin{figure*}[htb]
	\centering
	\includegraphics[height=.25\textheight]{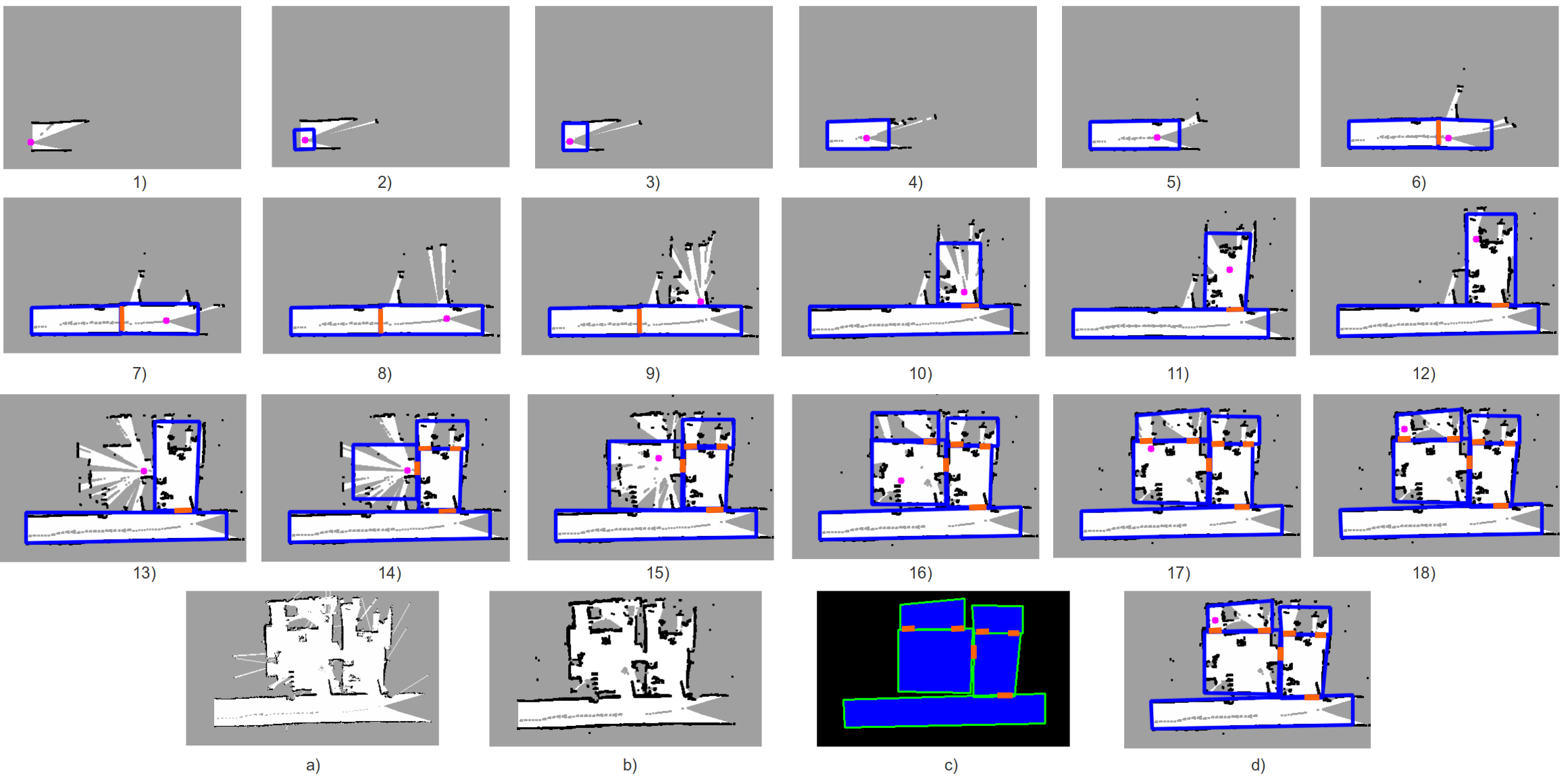}
    \caption{An example of the complete process of our online semantic exploration algorithm. 1) - 18): Snapshots of 18 steps of the process taken with different time intervals. The color coding is the same as that in Fig. \ref{figure:exp-final} d).}
    \label{figure:overall-process}
\end{figure*}

Fig. \ref{figure:overall-process} demonstrates an example of the complete process of our online semantic exploration algorithm. Parts 1) to 18) depict snapshots of 18 steps which are taken with different time intervals. Here, the violet point shows the robot position. In part 2), a new room is generated using the robot position and is put into the world through the sub-kernel ADD. In part 3) to 5), this room becomes bigger to better match the map through many successful DILATE. In part 6), 10) and 14) another three rooms are generated using the robot position and join the world by successful ADD. In part 10) two rooms are joined together by a successful MERGE. In part 16) one new room is generated independently on the robot position in the way that its neighbor room became first bigger through DILATE and was then decomposed into two rooms through a successful SPLIT. 

%Currently, we use a single-threaded implementation, where at each iteration only one sub-kernel is tested for the sampling. One biggest feature of the Markov chain is that the current state is only dependent on the previous one, therefore, it is theoretically possible to do multiple tests using different sub-kernels at each iteration, then only the successful test results are saved for the sampling. Using today's powerful off-the-shelf multi-core CPUs, this idea can be easily realized and should lead to a much less computation time.

\begin{figure}[htb]
	\centering
	\includegraphics[width=.72\columnwidth]{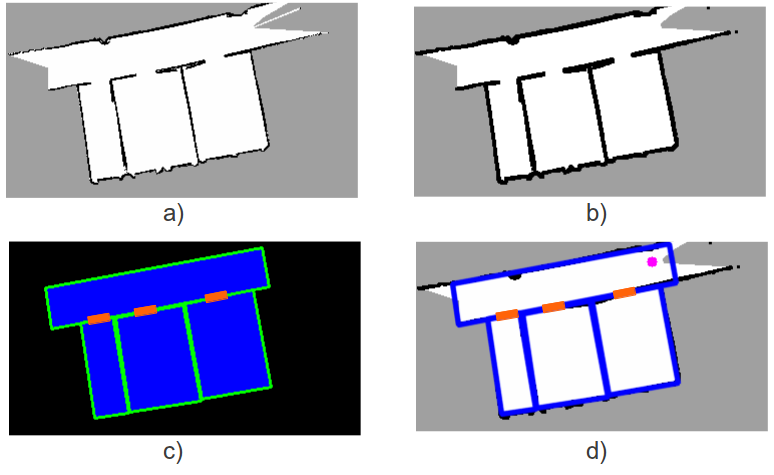}
    \caption{Final result for another indoor environment. The color coding is the same as in Fig. \ref{figure:exp-final}.}
    \label{figure:exp2}
\end{figure}

Final result of our semantic exploration algorithm for another indoor environment is shown in Fig. \ref{figure:exp2}. Since there is no furniture in this environment, it totally complies with our semantic model, thus, the result is better than that in Fig. \ref{figure:exp-final}.
%%%%%%%%%%%%%%%%%%%%%%%%%%%%%%%%%%%%%%%%%%%%%%%%%%%%%%%%%%%%%%%%%%%%%%%%%%%%%%%%
\section{Conclusions and future work}
This paper proposes a new approach for automatically extracting semantic information from preprocessed sensor data. We propose to do this by means of a probabilistic generative model and MCMC-based reasoning techniques. 
%Our work differs from previous semantic mapping approaches, that mostly use various classification methods in a bottom-up fashion to label either spatial regions or places based on context or that assign semantic labels directly to portions of the observations. 
We construct an abstracted semantic and top-down representation of the domain under the consideration: a classical indoor environment consisting of several rooms, that are connected by doorways. We use Bayesian reasoning to build this semantic map, so that it is aligned with the preprocessed sensor observations, that a robot made during an environment exploration and mapping stage. This introduces a bottom-up path into the approach and employs data driven discriminative environment feature detectors to analyze the continuous noisy sensor observations. 

%The semantic environment model that we generate, is structured similarly to a scene graph and is perfectly suited for any higher level reasoning and communication purposes. 

While we currently generate representations that more or less resemble a classical floor plan (including semantics however), the extension of our work to more functionally enhanced representations (e.g. differentiating several room types based on the context, adding other types of concepts like general objects or furniture) is obvious and will be pursued in the future. It is also straight forward to extend the concept towards 3D environment representations. 

%\addtolength{\textheight}{-3cm}   % This command serves to balance the column lengths
\section*{Acknowledgments}
This work is founded by the Institute for Advanced Study at the Technische Universit\"at M\"unchen.

%%%%%%%%%%%%%%%%%%%%%%%%%%%%%%%%%%%%%%%%%%%%%%%%%%%%%%%%%%%%%%%%%%%%%%%%%%%%%%%%
\bibliographystyle{plain}
\bibliography{semslam_icra2012}

%\begin{thebibliography}{99}
%\end{thebibliography}

\end{document}